# Brazilian License Plate Detection Using Histogram of Oriented Gradients and Sliding Windows

R. F. Prates[1], G. Cámara-Chávez[1], William R. Schwartz[2], and D. Menotti[1]

[1]Computing Department, University Federal of Ouro Preto, Ouro Preto, Brazil
[2]Computer Science Department, Univ. Federal of Minas Gerais, Belo Horizonte, Brazil

**ABSTRACT**

*Due to the increasingly need for automatic traffic monitoring, vehicle license plate detection is of high interest to perform automatic toll collection, traffic law enforcement, parking lot access control, among others. In this paper, a sliding window approach based on Histogram of Oriented Gradients (HOG) features is used for Brazilian license plate detection. This approach consists in scanning the whole image in a multiscale fashion such that the license plate is located precisely. The main contribution of this work consists in a deep study of the best setup for HOG descriptors on the detection of Brazilian license plates, in which HOG have never been applied before. We also demonstrate the reliability of this method ensured by a recall higher than 98% (with a precision higher than 78%) in a publicly available data set.*

## 1. INTRODUCTION

Intelligent Transportation Systems (ITS) are important tools for controlling traffic of vehicles in cities and highways. Nowadays, due to the increase in the number of vehicles, it has become an even more relevant research topic. These systems attempt to automatically identify vehicles in images or video sequences. One of the main components of an ITS is the Automatic License Plate Recognition (ALPR) system, which is present in several real-life applications such as automatic toll collection, traffic law enforcement, parking lot access control and road traffic monitoring. Another example of applicability of a license plate detection, which goes beyond ITS, is the privacy protection in Internet images containing vehicle license plate, such as in Google Street View, where is necessary to blur all license plate regions to maintain the privacy.

The ALPR systems are usually separated into four main modules: (1) image preprocessing, (2) license plate detection/localization, (3) character segmentation and (4) character recognition [1]. License plate detection is seen as the most challenge among these modules since it has to deal with a wide range of environmental conditions and license plate variations. Moreover, the detection results impact in the accuracy of the entire system. In spite of the existence of several commercial systems, license plate detection in open environments is still a challenge task.

License plate detection methods aim at finding rectangular regions in the image where license plates are correctly located. This task can be associated with techniques coming from several areas such as pattern recognition, image processing and computer vision. Existing methods explore one or a combination of license plate characteristics such as: (1) rectangular format with a known aspect ratio (depending on the image capture perspective); (2) distinguished color; and (3) rich edge texture due to the variation in color between characters and background.

Histogram of Oriented Gradients (HOG), a feature descriptor widely employed on several domains [2, 3, 4], characterizes objects through their shapes [2]. Since license plates are well represented by their shape, HOG is as excellent choice for license plate detection. HOG descriptors have already been successfully applied to the license plate detection problem showing their discriminative capacity of license plates between other descriptors [5, 6]. However, each country has license plates with specific characteristics, becoming necessary different characterization for each case. In Figure 1, it is shown the wide range of template and language variations in different countries around the world. It is important to note that there are different plate standards even in the same country due to different purposes accord to specific activity, e.g., public service vehicles, car rental, among others.

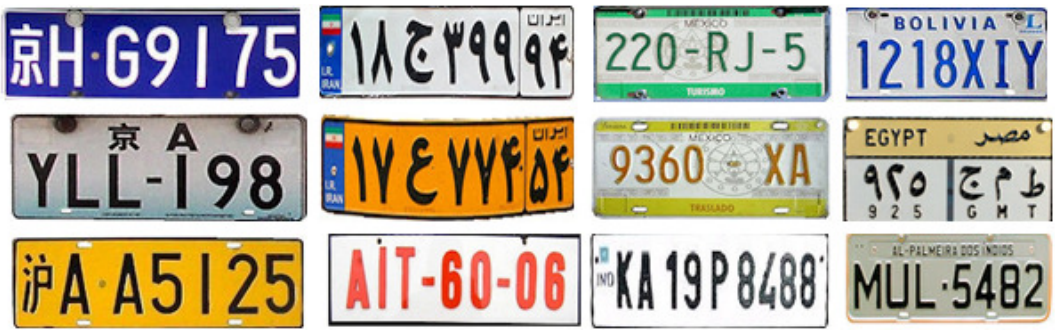

Fig. 1. License plates from different countries around the world.

In this paper, a license plate localization method based on a sliding window approach using HOG descriptors and a linear SVM classifier used on a multiscale setup is employed on Brazilian license plates (Figure 2 shows a schematic representation of our proposed method with its main modules). In order to speed-up the process, the Integral Histogram representation proposed in [7] is used. To the best of our knowledge, the evaluation performance of HOG descriptor in Brazilian license plate detection has never been done before, although it has been explored for various countries, e.g., China [6, 5] and EUA [8].

We can highlight our main contributions as the followings:

1. Different HOG setups are evaluated aiming at identifying the specific configuration that best performs in Brazilian license plates.
2. A multiscale sliding window detection algorithm using HOG descriptors is proposed and the detection effectiveness is analyzed;
3. A comparison is performed between the sliding window approach based on HOG features and the edge-based approach proposed in [9].

The remainder of this paper is organized as follows. Section II describes the related work and Section III describes the proposed sliding window approach using HOG features that searches for plates in multiscales. Section IV presents the databases used in our experiments and the results obtained by evaluating the HOG features parameters. Finally conclusions are pointed out in Section V.

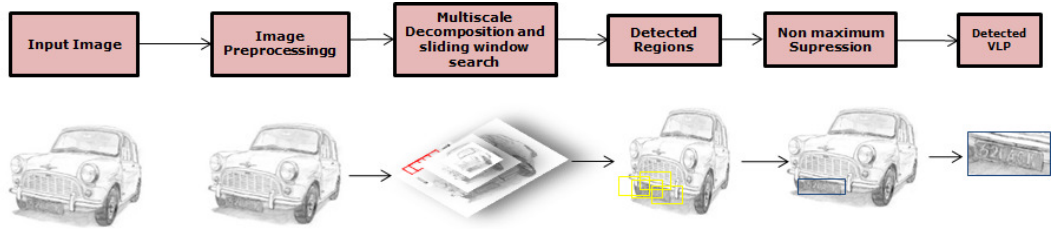

Fig. 2. Main modules of a method using sliding window and HOG descriptor in a multiscale way.

## 2. RELATED WORK

Some license plate detection methods employ edge information, focusing mainly vertical edges, extracted for instance by Sobel filters, once the car bumper shows rich horizontal edge features capable of confusing the algorithms [1]. In [10], the authors explore the vertical edge gradients and an image enhancement preprocessing based on the intensity variance of a region is applied to highlight the gradients in license plates. Some unwanted edges are further removed using the edge length. Then, a sliding window scans the whole image counting the total number of edge points in each delimited region and windows with high percentage of edge points are considered as a license plate. A drawback of that method is that the enhancement step highlights others regions besides the one of license plate. To surmount this problem, in [11], the authors proposed a new algorithm for image enhancement based on edge density. Since license plates regions exhibit strong vertical edge densities, this information can be used instead of intensity variance for contrast stretch in regions similar to license plates. A match filter based on mixture of Gaussians is employed to the edge density image to localize the license plate regions. These are then filtered using morphological operations to eliminate false positives. The effectiveness (precision/recall) of edge based approaches is usually low in complex images/scenes due to their high sensitivity to noise [6]. On the other hand they are able to achieve high detection rates when combined with morphological operations and geometric cues [12,13,14].

Hough transform is another method that is frequently applied to license plate detection problem. This method works well when the boundaries of license plates are clear and the images have been taken in a closed environment. The computational complexity associated with Hough transform algorithm is a strong drawback of this approach [1].

Aiming at measuring component area and aspect ratio, Connected Component Analysis (CCA) is applied to license plate extraction in [15]. Using CCA information, they were able to find the best plate candidates, reaching 96.62% of correct extraction and 1.77% of false alarms in a four hour low resolution videos.

Color features are also used in license plate detection, even though it depends on stable light conditions and can easily confuse the car body with the license plate due the color similarities [1]. In [16], the license plate detection is formulated as a visual matching problem. For each character, it is computed the SIFT features in the character region and the Principal Visual Word representation is built using unsupervised clustering. The assumption is that some characters can be seen as duplication of others being well represented by Bag of Visual Words models. In order to avoid background noise, the authors prefer to apply only the Principal Visual Words. Some geometric cues for each detected character are extracted to filter some false positive regions and identify the license plate size. Since they are matching characters, the recognition problem is partially addressed. This method does not work well in low resolution images and with severe perspective distortion since SIFT features are unable to be extract in this scenario and so the license plate location fails.

Sliding window techniques can be applied with different types of descriptors. In [17], a discrete wavelet transform is employed and the vertical and horizontal edges are computed from subband details, i.e., High-Low (HL) and Low-High (LH) subbands, respectively. A sliding window scans the whole HL image from top-left to bottom-right computing a histogram of vertical lines in order to eliminate vertical lines with height less than the average height of the region. A similar method is used to remove noise in LH subband. HL is then applied to extract some intensity transition features, while the LH is employed to verify if there are horizontal lines around the detected regions.

In [18], the author presented for the first time a combination of weak classifier for the license plate detection problem. The AdaBoost algorithm was employed to automatically construct a strong classifier from 100 weak classifiers. He proposed to use a sliding window technique to scan the image extracting Haar like features and locating license plate regions. The approach yields a detection rate of 95.6% and an error rate of 5.7%. Nonetheless this technique is highly dependent on the license plate size and has elevated computational cost since it does not use a cascade of classifiers. In [19], the authors also used an AdaBoost algorithm to select weak classifiers that consist in local Haar-features. Aiming at deal with different license plates size, they also employed a multiscale approach with detection window varying from 48×16 to 240×80 pixels. A cascade classifier is used to speed-up the detection algorithm with global features in the two first layers that are able to discard early more than 70% of the negative windows.

All AdaBoost methods mentioned before are based on the discrete AdaBoost (DAB) algorithm that is not well suitable in the literature due its low convergence and degradation. Moreover, it is empirically proved that the Gentle AdaBoost (GAB) performs better than DAB in the license plate detection problem [20]. In addition, in that work they proposed a cascade of 16 strong classifier detectors using Haar like features selected by GAB algorithm. The disadvantage of this approach is the high false positive rate that requires a post-processing step. In [6], the authors present a detection algorithm using GAB and Haar like features. However, unlike in [20], they use the gradient image instead of grayscale, since it is more robust for processing different plate colors. In order to eliminate false positive detections, some heuristics based on vertical edge projection and connect component analysis are applied along with a voted-based method using a SVM classifier. Different features have been analyzed showing that HOG is the most suitable for false positive reduction.

The method proposed in [21] is also based in two learning methods, where the first one is a GAB and the second a SVM classifier. However they applied a SIFT descriptor instead of HOG, and addressed the license plate detection problem as character detection, using 36 combined AdaBoost classifiers, one for each alphanumerical character. Another similar approach to the one in [6] is presented in [5], in which a GAB followed by a SVM classifier is employed in the license plate detection using Haar like features in an initial stage and further filtering the detected regions by applying HOG features. In that work, no heuristics is used before SVM classification becoming a more general method compared to [6]. Cascade frameworks usually have the drawback of dealing with high false positive rates. Usually increasing the number of layers in the cascade, decreases the false positive rate. Nonetheless, it is not a usual option due to the increasing in computational time and the decrease in the detection rate [5].

## 3. PROPOSED APPROACH

The applicability of HOG in license plate detection can be assured by recent works where false positive regions with texture like license plates could be rejected employing this feature to feed a linear SVM classifier [5, 6]. Moreover, previous works have shown that HOG can be successfully applied to different detection problems reaching high recall rates (detection) with high precision rates (few false positives) [3, 4].

To the best of our knowledge there is no works exploring HOG features in the detection of Brazilian license plates. Other works based on HOG to the license plate detection only explore its capability in a false positive filtering stage, which differs from our work. In addition, due the variation in the distance between the acquisition system and the vehicle, the problem has to be tackled at different scales. In this work, we use a sliding window approach based on Histogram of Oriented Gradients (HOG) features for Brazilian license plate detection. Details of this method are presented in the next sections.

### A. Histogram of Oriented Gradients (HOG)

The HOG descriptor was introduced by Dalal & Trigs in [2] based on the assumption that objects can be well represent by its shape and are locally computed in a dense grid. To obtain such discriminant information, the image is divided into cells and for each cell is computed a histogram of oriented gradients. Each pixel within the cell contributes with a weighted vote for an oriented histogram based on the values of the computed gradients. Horizontal and vertical gradients using Sobel 1-D are computed as follows.

$$dx = I(x+1, y) - I(x-1, y) \tag{1}$$
$$dy = I(x, y+1) - I(x, y-1) \tag{2}$$

From these gradients, the magnitude and orientation, respectively, are computed as

$$m(x, y) = \sqrt{dx^2 + dy^2} \tag{3}$$
$$\theta(x, y) = \arctan\left(\frac{dy}{dx}\right) \tag{4}$$

in which *I(x,y)* corresponds to the intensity value of an image at position *x* and *y*. In order to obtain a better invariance to illumination, these cells are organized in blocks and for each block normalization is conducted. In [2], the authors show that L2-norm is the most suitable for pedestrian detection (however, here L1-norm is adopted due to efficiency), i.e.,

$$v' = \frac{v}{\sqrt{\|v\|_2^2 + \epsilon^2}} \tag{5}$$

in which *v* is a non-normalized vector containing all histograms in a given block and ε is a small constant. Neighbouring blocks can be superposed in such a way that one cell can be present in different blocks and contribute distinctly in each histogram due to the normalization. Moreover, a Gaussian function is commonly applied to in order to attribute less weight for pixels near to borders of blocks than to center pixels. However, in [4], the authors improved their results in character detection using a Gaussian function of cells instead of blocks. With the magnitude and orientation of each pixel, a histogram of oriented gradients is then built such that the number of

International Journal of Computer Science & Information Technology (IJCSIT) Vol 5, No 6, December 2013

bins and range of values for $\theta$ that are assigned for each bin are parameters that highly depends on the application. In [2], $\theta$ is setup varying from 0 to $\pi$ in a set of 9 bins, while in [4] the values of $\theta$ vary from 0 to $2\pi$ in a set of 16 bins. Figure 3 shows a schematic representation of the HOG computation process, where each cell has 8 bins.

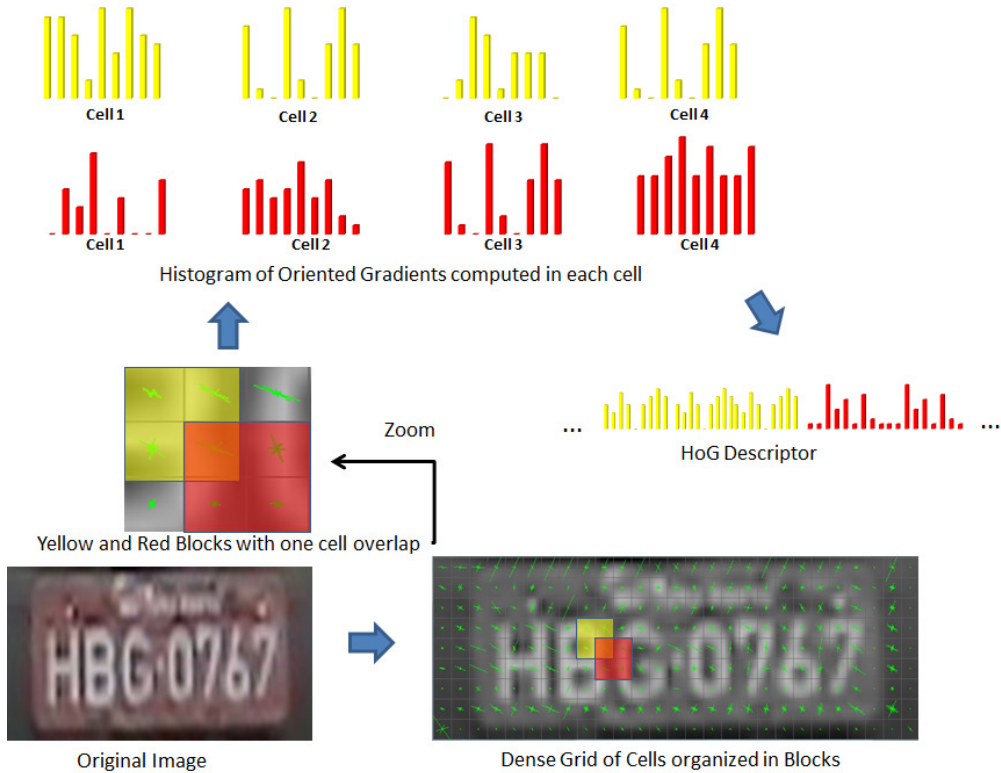

Fig. 3. Schematic representation of HOG descriptors extraction.

In order to detect objects in a wide range of scales, our approach builds a pyramid of images with a certain scale factor between two consecutive images. These images are then fully scanned with a fixed size window from top-left to bottom-right using a specific stride (distance between two consecutive windows) in a sliding window approach. For each of these regions, the HOG features are extracted and a classifier (in this case, a linear SVM) is used to determine the likelihood of such region being the object of interest. Figure 4 illustrates the process of sliding window in the pyramid of images.

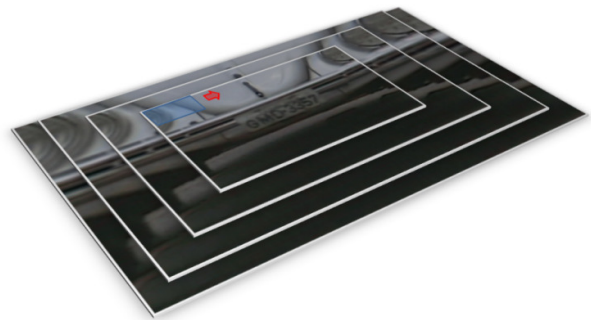

Fig. 4. Pyramid of images in different scales.

## 4. EXPERIMENTS

The experiments were performed on a Pentium Core i7, 2.10 GHz with 8 GB of RAM. Our implementation is on C++ using OpenCV 2.4.6 library and the source code is available in [25]. To perform the evaluation, we first describe the database used in the experiments and the evaluation metrics. Then, as the performance of a sliding window approach with HOG descriptors highly depends on the correct tune of various parameters, as discussed in [2], a detailed discussion of HOG parameters and an evaluation of HOG performance in the context of Brazilian license plate detection is presented. Finally, we present the results and compare them to the method described in [9] (since they provided their source code, such that a fully fair comparison could be carried out).

**A. Databases and Evaluation Metrics**

The tune of parameters and evaluation of the performance was carried out using a Brazilian publicly available database with 377 images (800 ×600 pixels) [9]. This database presents several challenges such as presence of fences, texts, pedestrians, regions with license plate like textures as roofs and other vehicle parts. The size of the plates also varies in a wide range since the distance between cars and acquisition system is not fixed. Figure 5 illustrates some samples considered in our tests. Besides original images, this database also let available images already manually labelled with rectangle bounding boxes as highlights Figure 5, allowing a reliable comparison of results.

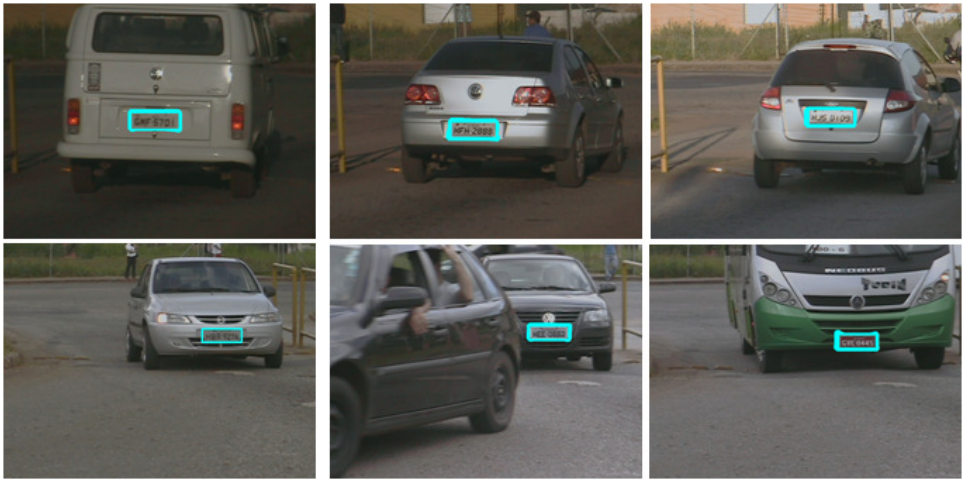

Fig. 5. Examples of the images from the database used in this work.

To perform the evaluation, we applied a protocol similar to the one used in [6] in which the match mp between two rectangular regions is computed as the intersection between these regions divided by their union. More specifically, the best match $m(r;R)$ for a rectangle $r$ to a set of rectangles $r_0 \epsilon R$ is defined as,

$$m(r; R) = max\{m_p(r; r_0)|r_0 \in R\} \qquad (6)$$

After performing the matching, the recall and precision, defined in Equations 7 and 8, respectively, are computed. In these equations, $T$ and $E$ represent the set of ground-truth and resulting rectangles, respectively, and $m_t(a,b)$, in which $a$ and $b$ represent two rectangular regions, is assigned to one if $m(a,b)$ is above a threshold $t$ and zero otherwise.

$$Recall = \frac{\sum_{r \in T} m_t(r, E)}{|T|} \qquad (7)$$

$$Precision = \frac{\sum_{r \in E} m_t(r, E)}{|E|} \qquad (8)$$

**B. Parameter Calibration**

The entire dataset, composed of 377 images, was first split in two sets: training and test. The training set is formed by 77 images and the test by 300. All parameter adjustments were conducted in the training set using a 5-fold cross validation. Aiming at obtaining the best configuration for Brazilian license plate detection problem, different HOG configurations were evaluated.

The protocol applied to the parameter calibration can be divided in four modules: (1) analysis of geometric characteristics of the license plates, including width, height, and aspect ratio to define the number of scales, size of sliding window and aspect ratio to be employed in the calibration tests; (2) evaluation of different HOG configurations performed by varying the number of blocks and cells to find the best combination of precision and recall; (3) adjustment of sliding windows to find the best stride between two consecutive windows; (4) evaluation regarding the influence of the background addition in the sliding window.

Histograms of height and width were built aiming at evaluating the variation of license plates size presents in the database. This is crucial information since it indicates the aspect ratio and the range of scales present in the database. Figure 6(a) presents the histogram of the width of the license plates. The mean and standard deviation values of the width of plates are $99.5 \pm 12.64$ pixels. In Figure 6 (b), it is shown the histogram of the height of the license plates, and the figures are $32.5 \pm 4.68$ pixels. Finally, the aspect ratio, calculated using the width of license plates divided by the height, is $3.1 \pm 0.19$. How it was expected, the aspect ratio obtained is very close to the specified in the Brazilian Legislation [26], that determines dimensions for license plates with 130 and 400 mm in vertical and horizontal, respectively. Based on these geometric cues, we chose to fix the sliding window size to 90×30 pixels, since this size is according to the mean aspect ratio and also is near to the mean of width and height sizes of the license plates.

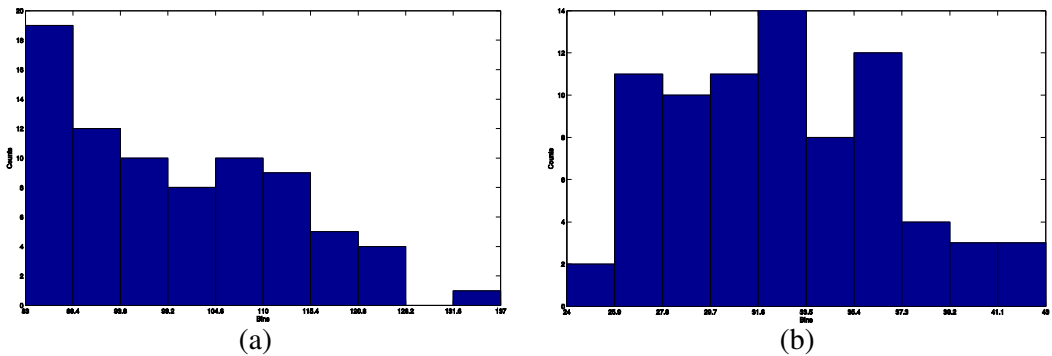

Fig. 6. Brazilian license plate size statistics: (a) Histogram of the width. (b) Histogram of the height.

In [2], the authors show that adding background information to the sliding window may increase the effectiveness of detection. Therefore, to capture better background information, a study of the number of pixels that can be added to the 90×30 window is conducted. For this purpose, we define some initial configuration for HOG descriptors and the number of scales to be able to evaluate the influence on these last two parameters for different amount of background addition. For this study, we use cells of 6×6 pixels organized in blocks of 2×2 cells in a multiscale sliding window detection with 11 scales.

Figure 7 shows that on the one hand, removing pixels from the license plate drastically reduces the performance. However, on the other hand, the addition of pixels to the sliding window increases the effectiveness until a certain number of pixels, in our case 9×3 pixels. According to the results, we selected to add 9 and 3 pixels to the sliding window (9 pixels were added to both horizontal sides of the license plate and 3 pixels to both vertical sides), resulting in a sliding window of 108×36 pixels instead of the initial 90×30 pixels.

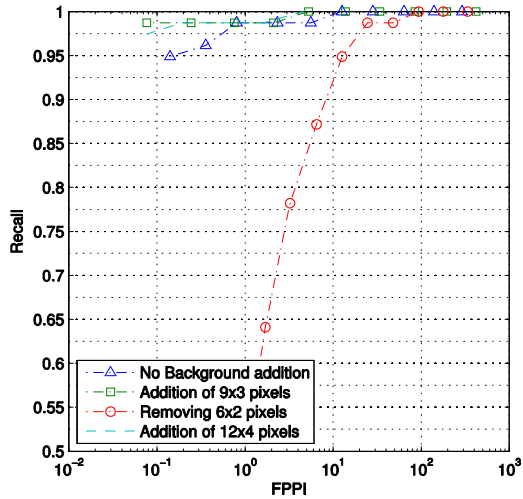

Fig. 7. Different background configurations.

Fixing a threshold for the SVM classifier which yields 1 false positive per image (FPPI) and employing 11 scales, we perform an evaluation of different block configurations and cells for the HOG descriptor.

Figure 8(a) presents the result of these experiments in terms of recall, while in Figure 8(b) the precision is shown. Considering the trade-off for these two metrics, we opt to use cells of 4×4 pixels organized in blocks of 2×2 cells, once with this configuration we reached the best precision with a fair recall.

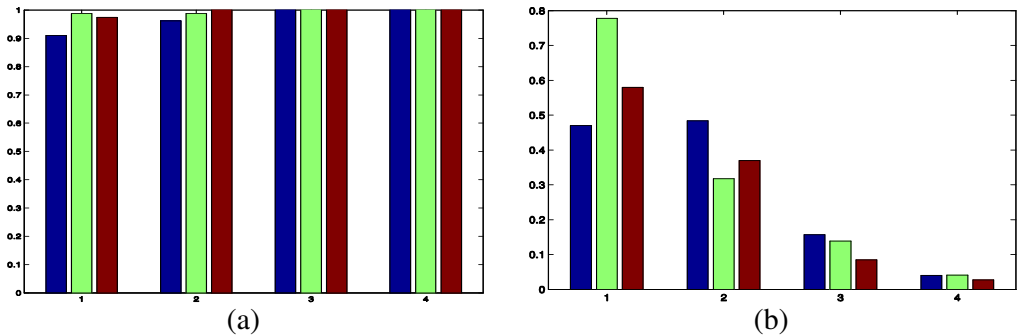

Fig. 8. HOG setup: (a) Recall and (b) Precision for different HOG configurations, where 1, 2, 3 and 4 stand for 4x4, 6x6 9x9, and 12x12 pixel cells, respectively, and blue, green and red columns stand for 1x1, 2x2, and 3x3 cell block configuration, respectively.

Here, we analyze the parameters that are directly associated to the multiscale approach – the number of scales and the stride between adjacent windows. The number of scales is an important parameter that affects both efficiency and accuracy.

In Figure 9(a), the results evaluating the number of scales using a HOG descriptor of 4×4 pixel cells organized in 2×2 cell blocks. We can see that with 11 scales, the system performs very well, with 97% of recall at 0.1 FPPI - note that we used 11 scales as maximum to avoid prohibitive computational cost. In addition, Figure 9(b) presents the performance of the system for different strides between two adjacent windows. In this analysis, we search for the higher stride that can be applied without reducing significantly the detection rate. Using this criterion, 9 pixels of stride was chosen.

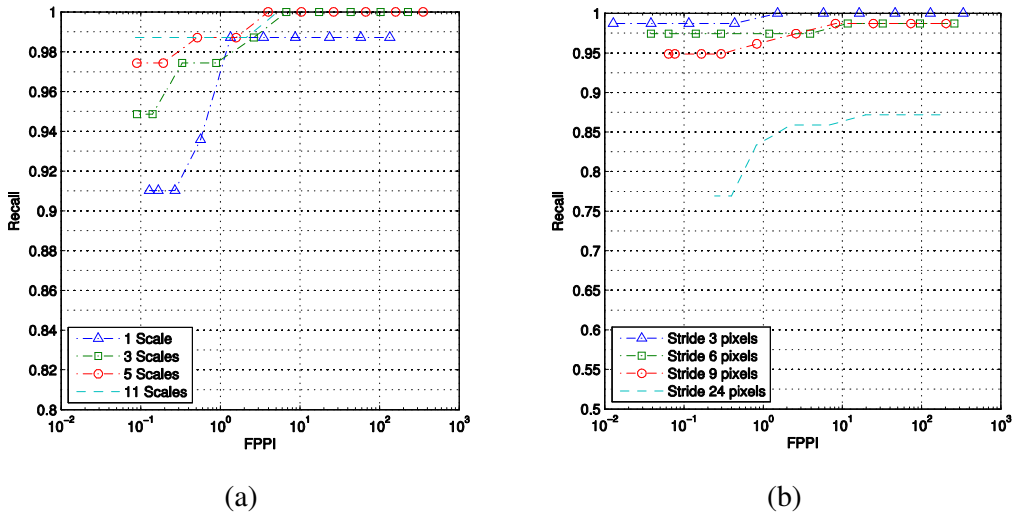

(a)　　　　　　　　　　　　　(b)

Fig. 9. System performance: (a) at different number of scales; (b) at different strides.

## C. Experimental Setup

As explained in the earlier in this section, the parameters of the sliding window approach were adjusted using a training/validation set composed of about 20% (77) of images, and the remaining 80% (300) was used for test. It is important to remark that our approach is inferior to method proposed in [2], since we use L1-norm instead of L2-norm and, did not apply Gaussian mask and tri-linear interpolation for constructing the HOG descriptor for each block of cells. While fast to compute, Integral Histogram limits these steps in HOG computation, once they do not fit well [23]. Here, we summarize the values used for each parameter to present the ultimate results.

**Detection Window**: 108×36 pixel window (counting 9 and 3 pixels horizontally and vertically added at each direction) is employed for window size. **HOG descriptor**: The HOG descriptor configuration used is a 4×4 pixels cell grouped in blocks of 2×2 cells, generating a feature vector with 7,488 variables per detection window. L1-norm is employed to normalize the HOG histogram extracted from each block. **Sliding window approach**: The stride between two detection windows is 9 pixels in the sliding window search. The multiscale search is conducted with 11 scales. **SVM learning**: The SVM threshold is adjusted to reach a 1 False Positive per Image (FPPI) rate at maximum and the regularization parameter C is setup using cross-validation on the training set. For each database, a SVM classifier is separately trained. Negative samples were automatically extracted from the background (with no intersection to the labelled real license plates). Initially, 20 negative examples for each image were sampled. After with one round bootstrapping (adding harder examples to train the classifier, bringing down the false positive rates), 1000 more negatives samples were added to the training set.

### D. Experimental Results and Comparisons

Figure 10 shows examples of some incorrectly detected regions for the multiscale sliding window approach using HOG features and a hybrid edge detection method introduced in [9]. In this Figure, bounding boxes in red are displayed to represent the ground truth regions of the license plates, while the blue ones show the detected bounding boxes. The results present in the first row were obtained by using hybrid-based approach proposed in [9], while the bottom row shows detections of the multiscale sliding window method. These images show that both methods are sensitive to region with texture like license plates. However, as we can observe from these images, the hybrid-based approach is more sensitive to these regions by often detecting car headlight and texts regions present in the images.

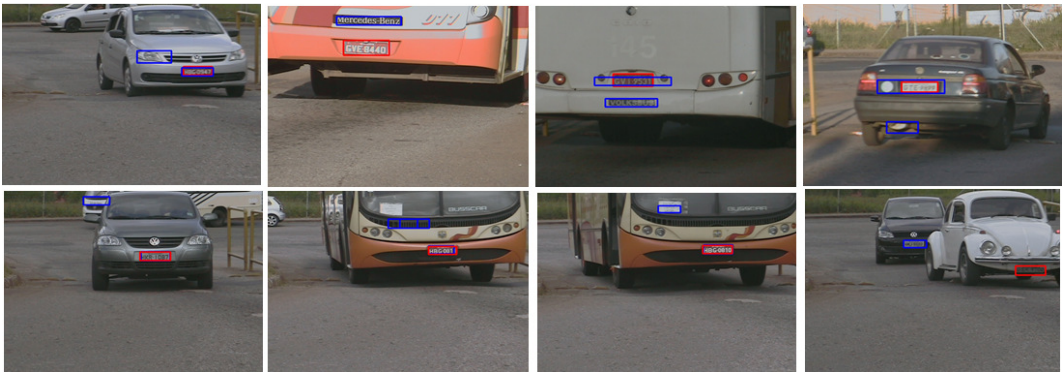

Fig. 10. Example of detections for hybrid [9] (top row) and sliding window using HOG features (bottom row) approaches.

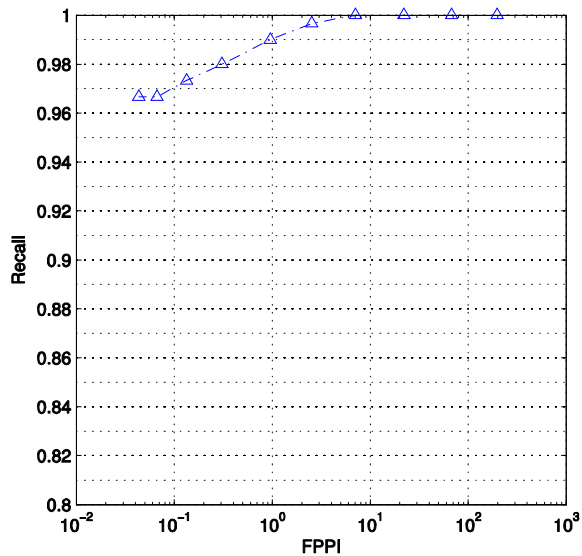

Fig. 11. DET Curve for the multiscale sliding windows approach using HOG features.

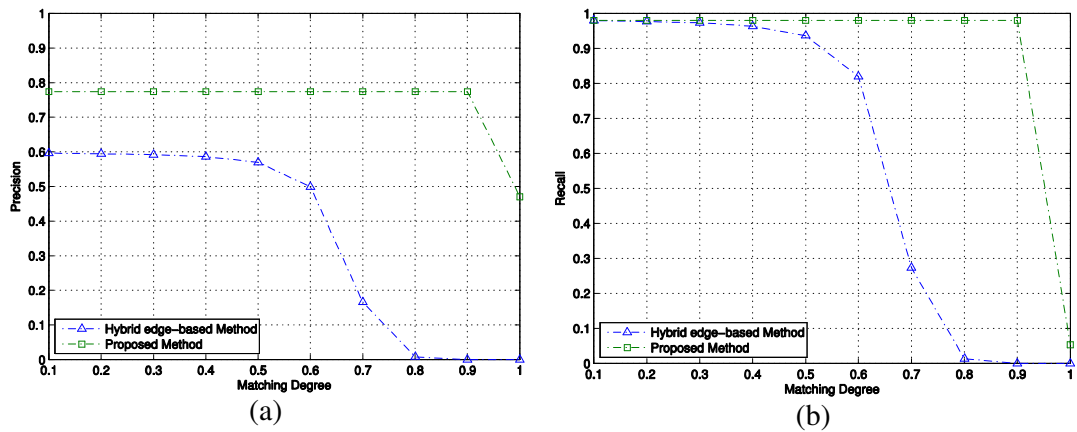

Fig. 12. Effectiveness versus matching degree: (a) Precision and (b) Recall. Curves in blue-triangle and green-squared represent the effectiveness of the Hybrid edge-based method and sliding window-HOG descriptor approach, respectively.

Figure 11 shows the detection error trade-off (DET) curve obtained using the multiscale sliding window with HOG descriptor approach in the Brazilian database. As we can see, the method using HOG features in a multiscale sliding window approach performs very well in this specific problem reaching promising detection rate of 99% at 1 FPPI and outperforms the hybrid-based method [9] that achieve 93.67% at 0.71 FPPI.

Figure 12(a) shows that the recall of the hybrid-based approach rapidly degraded as the matching degree requirements became higher. This behaviour is also observed in the Precision versus Matching degree curve. In this case, our method outperforms the hybrid-approach in all matching degree range.

Note that the multiscale sliding window approach is capable to obtain a better matching degree, which is critical for the further character recognition stage. In addition, observe that in both methods there is just one license plate labeled for image and in some situations more than one vehicle can be present in the image. In such case, the nearest license plate close to the acquired system (i.e., the digital camera) was considered as the target. However, such cases highly degrade the reported results, decreasing the detection rate (e.g., in Figure 10 bottom-right) or increasing the FPPI. It is important to note that a matching degree higher than 0.85 is considered as outstanding for further character recognition [9], and matching degrees smaller than 0.6 made very hard the task of full character recognition since at least one of the characters of the license plate should be missing in the located plate region.

## 5. CONCLUSIONS

To the best of our knowledge, the applicability of HOG descriptors in the Brazilian license plate detection problem is analyzed for the first time. Since license plates in different countries can present different template, size, color and language we firstly perform a tune of parameters, aiming at discovering the best configuration for HOG and multiscale sliding window approach with the minimum computational cost and maximum effectiveness. In a second stage we evaluated our method in publicly available database and the results were compared to a hybrid edge-based method proposed in [9] showing that the HOG based approach yields higher recall and precision rates. We also show that our method is more suitable for the further recognition stage, once its detection matching degree is very higher.

Despite the outstanding results, our implementation is still expensive for a real-time requirement of ALPR system. We credit these disadvantages to the implementation, which can be significantly accelerated using GPU for HOG descriptor computation and SVM classification in future works.

## ACKNOWLEDGEMENTS

This work was supported by the CNPq/MCT, CAPES/MEC, FAPEMIG, and FAPESP, Brazilian Government's research support agencies.

**Authors**

**Raphael Felipe de Carvalho Prates** received his Automation and Control Engineer degree from the Federal University of Ouro Preto (UFOP) in 2012, and now a days (2012) he is a Master student in Computer Science, UFOP. 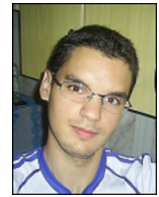

His research interests include image processing, pattern recognition, computer vision, and computer intelligence applied to process control**.**

**Guillermo Cámara Chávez** received the M.Sc. degree in computer science in 2002 from the University of São Paulo, São Paulo, Brazil, and the Ph.D. degree in computer science in 2007 from the Federal University of Minas Gerais, Belo Horizonte, Brazil and the Ph.D. in image and signal processing from the University of Cergy-Pontoise, Paris, France. He is currently an associate professor at Federal University of Ouro Preto since 2009. 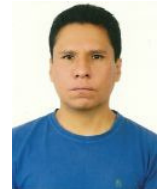

His research interest focus on video and image processing, computer vision, pattern recognition, and content-based information retrieval.

**William Robson Schwartz** is a Professor in the Department of Computer Science at the Federal University of Minas Gerais, Brazil. He received his BSc and MSc degrees in Computer Science from the Federal University of Paraná, Curitiba, Brazil in 2003 and 2005, respectively. He received his PhD degree in Computer Science from the University of Maryland, College Park, USA in 2010. Then, he spent one year in the Institute of Computing at the University of Campinas as a Postdoctoral researcher. 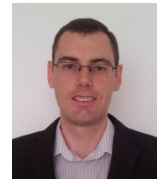

His research interests include Computer Vision, Surveillance, Forensics, and Biometrics, with focus on problems of face spoofing and recognition, human detection, and person re-identification. He has served as a Program Committee member for conferences such as IEEE International Conference on Automatic Face and Gesture Recognition (FG) and IEEE Workshop on the Applications of Computer Vision (WACV).

**David Menotti** received his Computer Engineering and Informatics Applied Master degrees from Pontifícia Universidade Católica do Paraná (PUCPR), Brazil, in 2001 and 2003, respectively. In 2008, he received his cotutelle PhD degree in Computer Science from both the Universidade Federal de Minas Gerais (UFMG), Belo Horizonte, Brazil and the Université Paris-Est/Groupe ESIEE, Paris, France. He is an associate professor at the Computing Department (DECOM), Universidade Federal de Ouro Preto (UFOP) since 2008. And currently, he is working as collaborator professor at Post-Graduate Program of Computer Science, Computer Science Departament (DCC), UFMG. 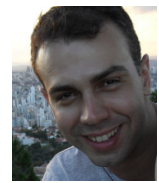

His research interests include image processing, pattern recognition, computer vision and algorithms for information retrieval systems.